\documentclass{uscthesis}

\usepackage[style=numeric]{biblatex}
\bibliography{references}

\usepackage{graphicx}
\graphicspath{ {./images/} }
\usepackage{multirow}
\usepackage{tabularx}
\usepackage{dsfont}

\usepackage{amsfonts}       
\usepackage{amsmath}

\newcommand{\topone}[1]{\textbf{\textcolor{black}{#1}}}

\definecolor{dg}{rgb}{0,0.694,0.298}
\definecolor{purple}{rgb}{0.4,0.176,0.569}
\definecolor{royalblue}{RGB}{65,105,225}
\usepackage{pifont}
\newcommand{\figref}[1]{Fig.~\ref{#1}}
\newcommand{\reqref}[1]{Eq.~\eqref{#1}}
\newcommand{\secref}[1]{Chapter ~\ref{#1}}
\newcommand{\tableref}[1]{Table~\ref{#1}}

\title{CNN-based Dendrite Core Detection from Microscopic Images of Directionally Solidified Ni-base Alloys}

\author{Xiaoguang}{Li}    

\degreedate{2022}                      

\otherdegrees{
Bachelor of Science\\
Zhengzhou University 2013\\ 
}                          

\degreename{Master of Science}     
\field{Computer Science and Engineering}          
\college{College of Engineering and Computing}  

\advisor {Dr.}{Song Wang}{Director of Thesis}  
\readera{Dr.}{Yan Tong}{Reader}     
\readerb{Dr.}{Ramtin Mohammadizand}{Reader}          
\readerc{Dr.}{Jeff Simmons}{Reader} 

\dean{Tracey L. Weldon}{Vice Provost and Dean of the Graduate School}   

\copyrightpage       

\abstract{0_abstract}  

\makeLoT               

\makeLoF               

\begin{document}

\chapter{Introduction}
\label{chapter:motivation}


    






    





\begin{figure}[h!]
  \includegraphics[width=0.8\textwidth]{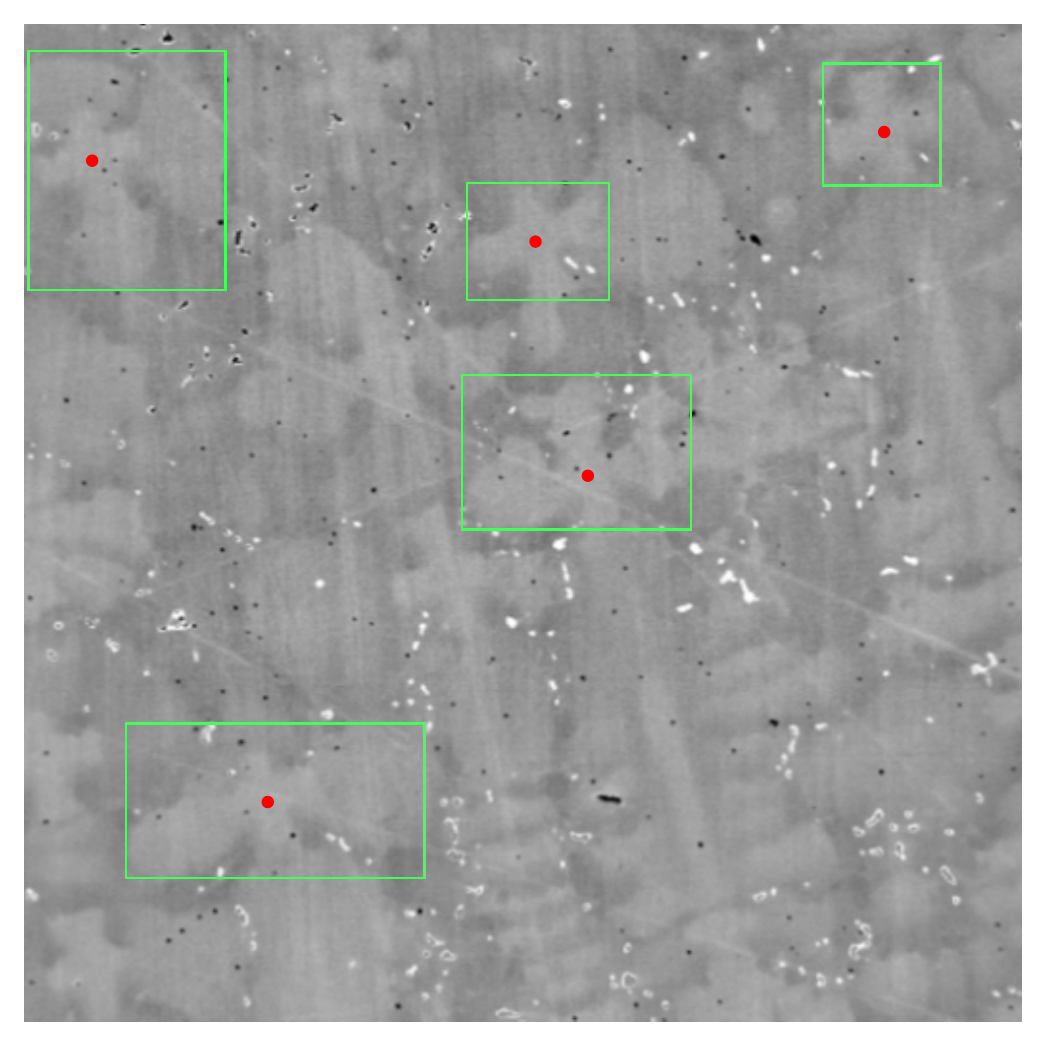}
  \centering
  \vspace{-25pt}
 \caption{An illustration of dendrite samples. The red points denote the dendrite cores.}
  \label{fig:dendrites}
\end{figure}
\begin{figure}[h!]
\centering
  \includegraphics[width=0.9\textwidth]{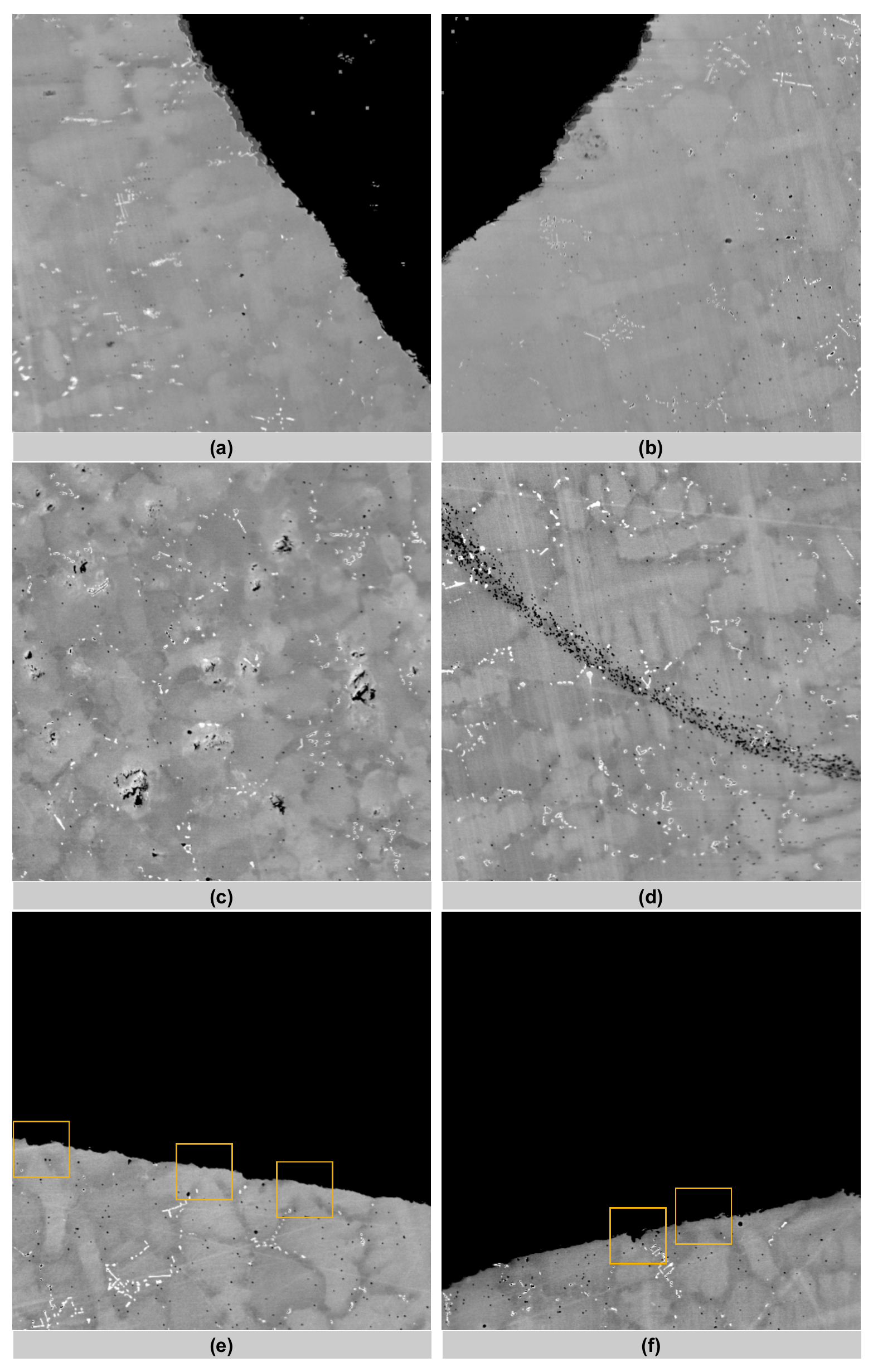}
  \vspace{-25pt}
 \caption{Examples of microscopic images with blurry and noise in (a)(b) and (c)(d) respectively and the incomplete dendrites in (e)(f).}
  \label{fig:hard_blury_noise}
\end{figure}

Dendrites are tree-like structures of crystals which are formed during some materials' solidification  \cite{takaki2013unexpected}. In the procedure of solidification, the morphology of dendrite can has a huge change which leads to a significant effect on the properties of materials \cite{sakane2018three}  \cite{gibbs2015three} \cite{daudin2017particle} \cite{li2004real}. For example, \cite{sun2013structural} claims that learning the growth mechanism of metal dendrites in the electrochemical procedure can extend battery life significantly and \cite{neumann2017general} mentions the study of dendrites structure is helpful to predict solidification defects. The dendrite core is the center point of the dendrite. Several dendrite samples are shown in \figref{fig:dendrites} where red points denote the dendrite cores. The information of dendrite core is very helpful for the material scientists to analyze the properties of materials. Therefore, detecting the dendrite core is a very important task in the material science field. Meanwhile, because of some special properties of the dendrites, this task is also very challenging. First, the microscopic images of the dendrites can be very blurred as shown in \figref{fig:hard_blury_noise} (a)(b), and it can be very difficult to locate the dendrite cores from the microscopic image. Second, besides the blurry, there are many different noises in the microscopic images such as the dark and bright points in \figref{fig:hard_blury_noise} (c) and the black spots in \figref{fig:hard_blury_noise} (d). These noises will increase the difficulty of locating the dendrites. Third, some of the dendrites in the microscopic image are incomplete such as the dendrites in the yellow boxes in \figref{fig:hard_blury_noise} (e)(f). The appearance of these incomplete dendrites is totally different from that of the complete ones and only the materials scientists with expertise can recognize them. As a result, locating the dendrites and annotating the dendrite cores are very time-consuming and expensive.

\begin{figure}[h!]
\centering
  \includegraphics[width=1\textwidth]{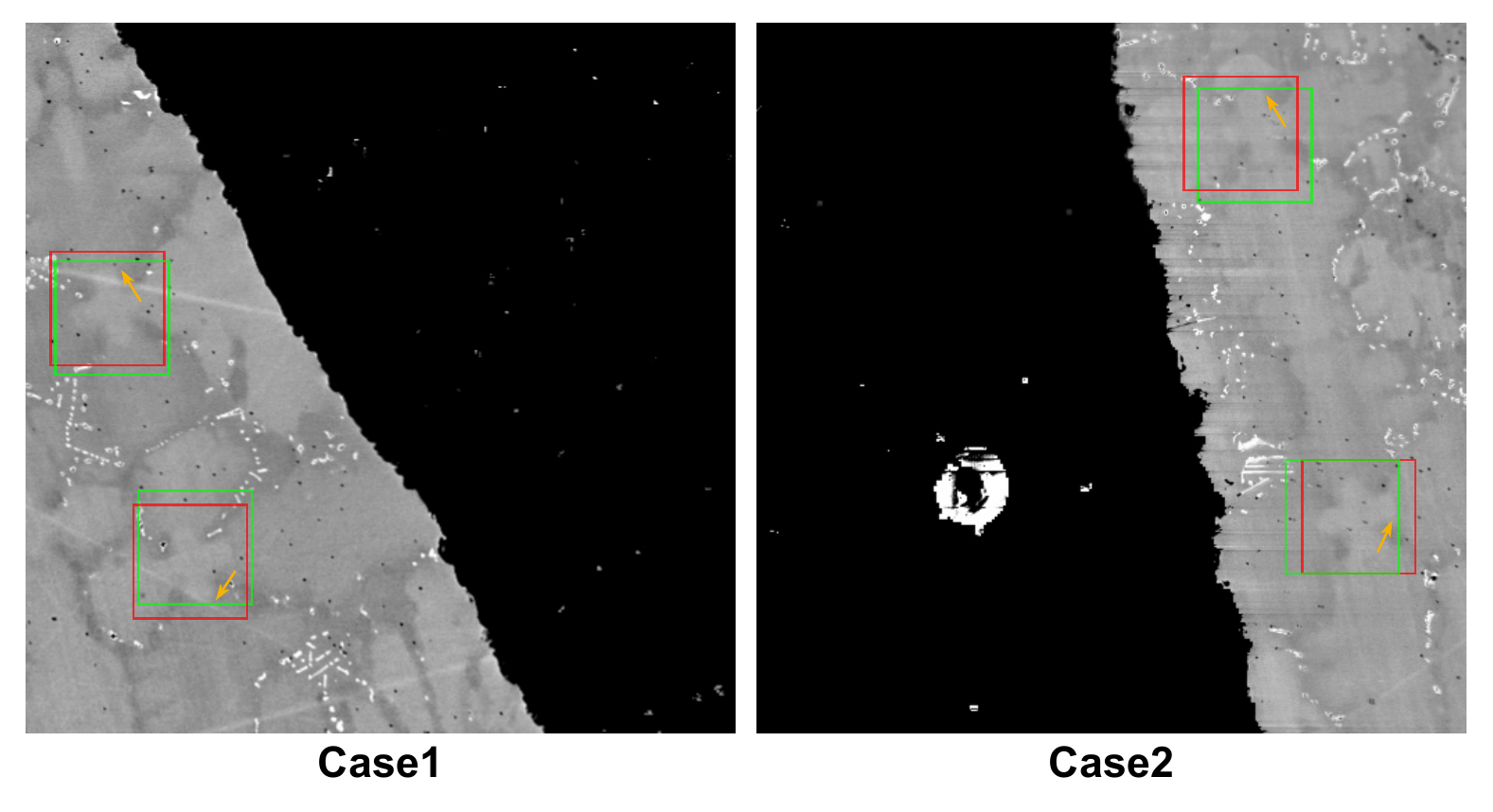}
  \vspace{-25pt}
  \caption{Visualization results of the regressing bounding-box based method. The red boxes denote the ground truth and the green boxes denote the predictions. The yellow arrows denote the edges that the predicted boxes try to fit.}
  \label{fig:boxe_edge}
\end{figure}

As Deep Neural Network (DNN) has achieved significant success in the natural image process, material scientists are exploiting DNN to solve the problems in their field. For example, \cite{zhou2016large} proposes to use DNN to reconstruct 3D objects by detecting 2D objects on each cross-section, and \cite{konopczynski2021deep} proposes to use DNN to perform X-ray CT segmentation task. However, there are few kinds of research focusing on dendrite core detection. Different from the typical detection problems in the computer vision field, detecting the dendrite core aims to detect a single point location instead of the bounding-box. Therefore, the existing regressing bounding-box based detection methods such as \cite{liu2016ssd}, \cite{redmon2016you}, \cite{ma2018arbitrary}, and \cite{ren2015faster} can not work well on this task. From \figref{fig:boxe_edge} we can find that the regressing bounding-box based methods try to tightly cover the dendrite instead of focusing on the dendrite core. Besides, the appearance of the dendrites varies a lot, therefore the calculated center point location based on the upper-left and lower-right corners of the bounding box is usually inaccurate when used as the estimate of the dendrite core. As for the key points detection algorithms, they also can not work well on this task because of the complex properties of the dendrites mentioned above. For example, the key point detection method in \cite{newell2016stacked}, which is designed for detecting human key points,  can not be used to detect the dendrite cores, even after increasing network complexity, due to the blurry or serious incompleteness of the dendrites.

Inspired by \cite{newell2016stacked}, In this work, we formulate the dendrite core detection problem as a segmentation task and propose a novel detection method to detect the dendrite core directly. Our whole pipeline contains three steps: Easy Sample Detection (ESD), Hard Sample Detection(HSD), and Hard Sample Refinement (HSR). Specifically, ESD and HSD focus on the easy samples and hard samples of dendrite cores respectively. Both of them employ the same Central Point Detection Network (CPDN) but not sharing parameters. To make HSD only focus on the feature of hard samples of dendrite cores, we destroy the structure of the easy samples of dendrites which are detected by ESD and force HSD to learn the feature of hard samples. HSR is a binary classifier which is used to filter out the false positive prediction of HSD.

Our main contribution is twofold.

\ding{182} We propose a novel detection method to detect the dendrite cores directly.

\ding{183} We conduct a series of experiments for exploiting the optimal crop size and crop intensity to destroy the structure of the easy samples of dendrites.

\chapter{Background}

\section{Convolutional Neural Network (CNN)}
The Convolutional Neural Network (CNN) is similar to the fully connected neural network. Both CNN and the fully connected neural network are comprised of neurons and can optimize their parameters through the learning process. However, compared with the fully connected neural network, CNN has several salient advantages. First, CNN is much more computationally efficient because of parameter sharing. Second, the sparsity of connection in CNN makes each output of the convolution layer only depend on a small number of inputs which efficiently prevent the overfitting. Third, CNN can keep the position information of the 2D images. Since the AlexNet \cite{krizhevsky2014one} is proposed in 2012, CNN has almost become the most popular deep learning architecture in the computer vision field. Usually, the whole CNN contains several components such as convolution operation, pooling, activation function, stride, and padding. We will elaborate on each of these components in the following paragraphs. 

\subsection{Convolution}
\begin{figure}[h!]
  \centering
  \includegraphics[width=1\textwidth]{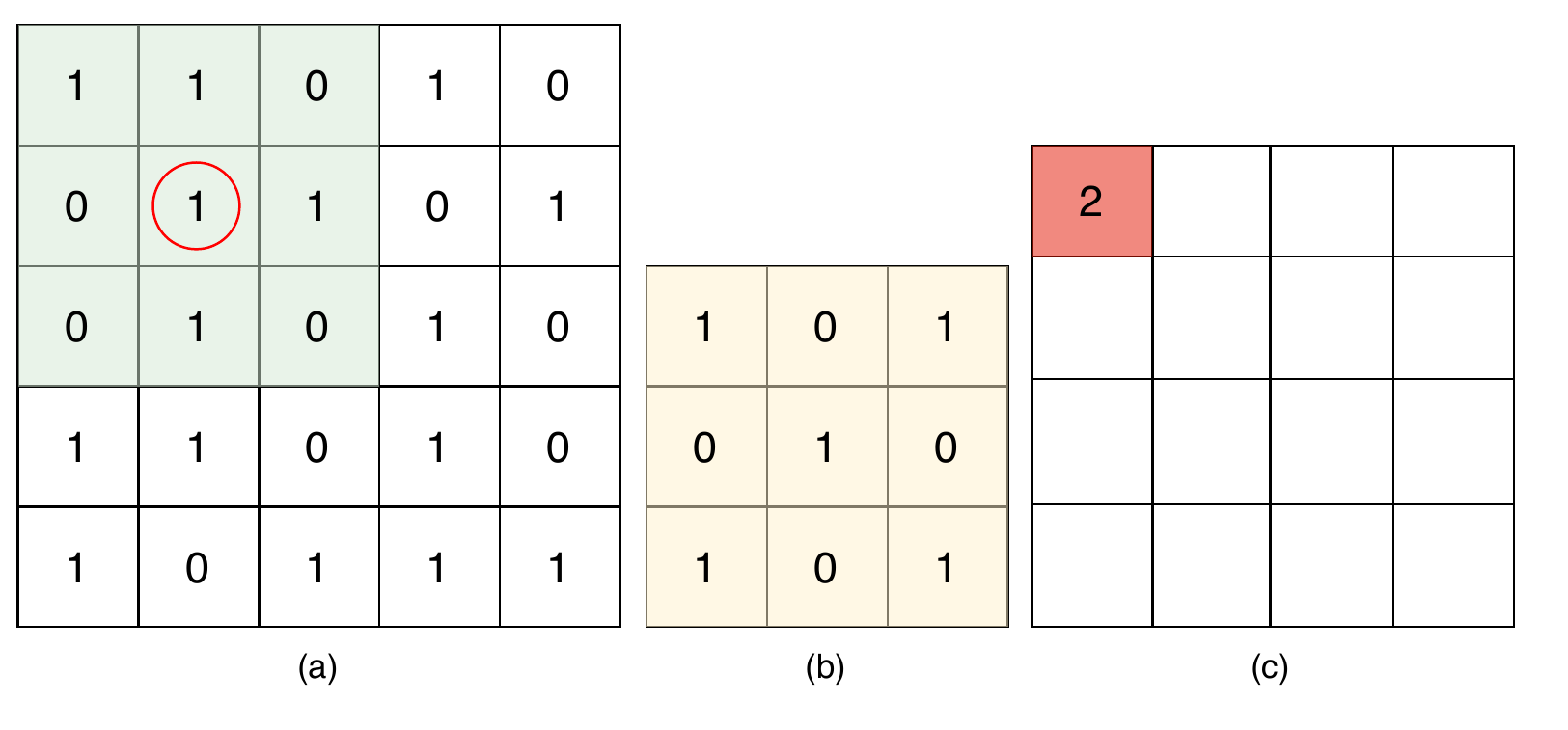}
  \vspace{-25pt}
  \caption{(a) denotes the input vector, (b) denotes the filter or kernel, and (c) denotes the result of convolution operation at the red circle location of the input vector.}
  \label{fig:cnn}
\end{figure}

The convolution operation is the most important component in CNN. Specifically, the convolution operation is a mathematical operation between the input vector and the corresponding filter or kernel.  As shown in \figref{fig:cnn}, it performs the convolution operation at the red circle location of the input vector. It makes element-wise multiplication and sums the result. In this case, the kernel size is 3*3, therefore, it will take a 3*3 area at each location of the input vector. In practice, in each convolution layer, it chooses many different kernels to do the convolution operation because each of these kernels can extract different feature information from the input vector. Based on the previous study, the shallow convolution layers of CNN can extract low-level features from the input vector such as edges, and the deeper layers can extract possible objects such as faces or even more complex features.

\subsection{Activation function, stride, and padding}
\begin{figure}[h!]
  \centering
  \includegraphics[width=0.7\textwidth]{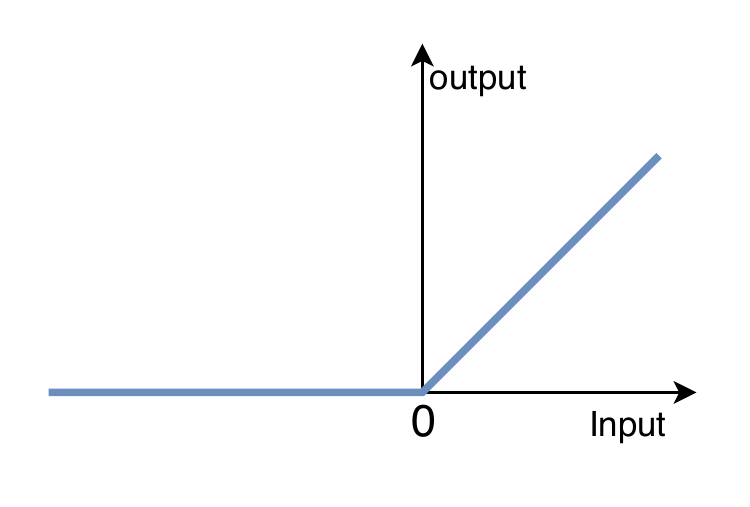}
  \vspace{-25pt}
  \caption{ReLU function.}
  \label{fig:relu}
\end{figure}

The convolution operation is linear. In order to make CNN more powerful, it passes the result of the convolution operation through a non-linear function such as ReLU in \figref{fig:relu}.  The ReLU function drops the values smaller than zero. The stride specifies the distance it moves the filter or kernel at each step. We can choose a big stride if we want to have less overlap between adjacent convolution operations. In order to keep the output feature having the same dimension as the input feature, we also pad zeros around the input vector.

\subsection{Pooling}
\begin{figure}[h!]
  \centering
  \includegraphics[width=0.7\textwidth]{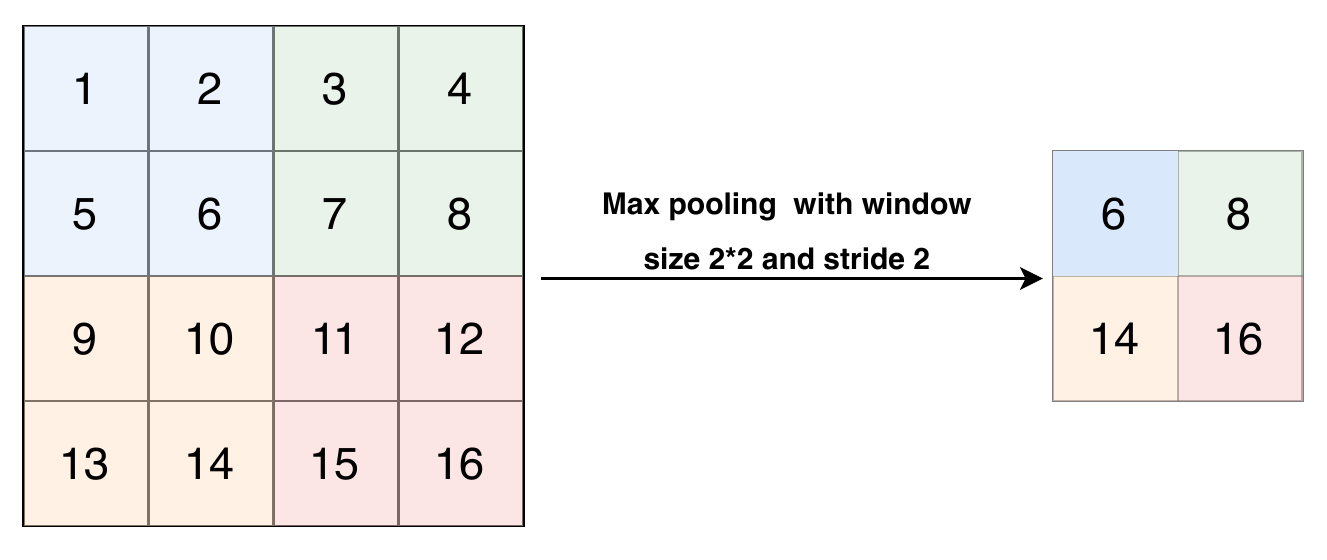}
  \vspace{-25pt}
  \caption{Max pooling.}
  \label{fig:pooling}
\end{figure}

The pooling operation reduces the feature’s dimension which can save the training time and prevent overfitting. There are no trainable parameters in the pooling operation. For the max-pooling in \figref{fig:pooling}, it keeps the maximum value around the 2*2 window. In this case, the max-pooling downsamples the feature dimension to 2*2 from 4*4.

\section{Residual learning framework}

\begin{figure}[h!]
  \centering
  \includegraphics[width=0.7\textwidth]{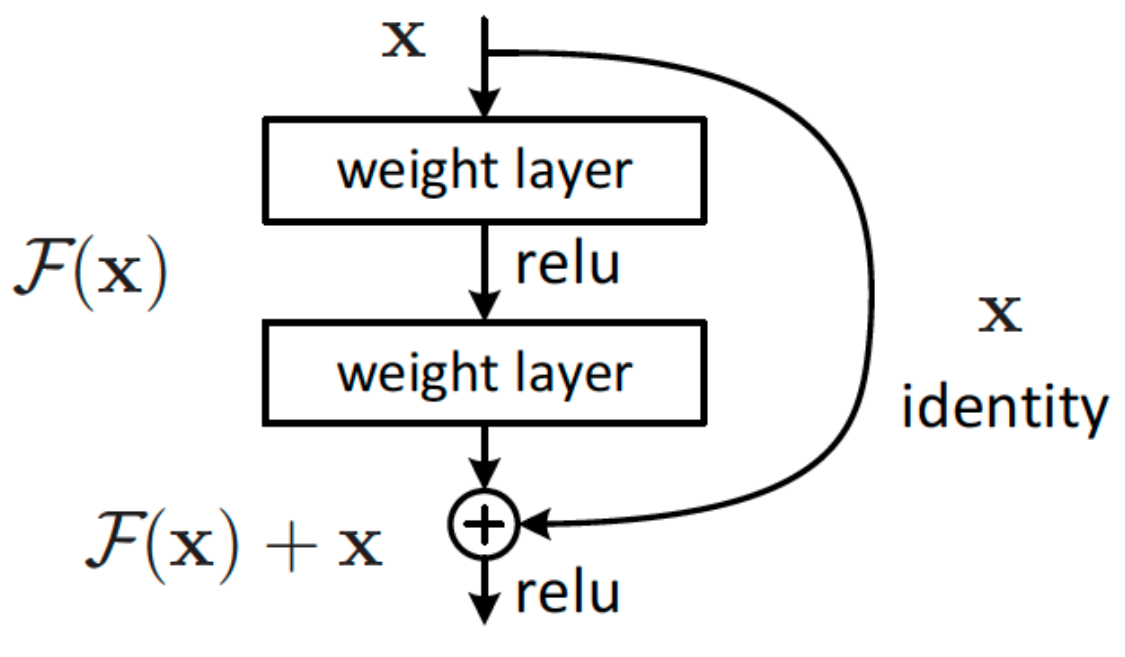}
  \vspace{-25pt}
  \caption{Residual learning block. The weight layer denotes the convolution layer. The figure comes from \cite{he2016deep}.}
  \label{fig:residual}
\end{figure}

It is difficult to train a deep neural network when it contains a lot of layers. As the depth of the deep neural network increases, the accuracy degrades quickly. In order to solve the degradation problem, \cite{he2016deep} propose the “deep residual learning framework”. $\mathbf{x}$ and $\mathbf{F}(\mathbf{x})$ in \figref{fig:residual} denote the input feature and the residual respectively. $\mathbf{H}(\mathbf{x})$ in \reqref{eq:residual}
\begin{align}
\label{eq:residual}
\mathbf{F}(\mathbf{x}) = \mathbf{H}(\mathbf{x}) - \mathbf{x}
\end{align}
denotes the desired underlying mapping. Instead of making the layers fit $\mathbf{H}(\mathbf{x})$ directly, the residual learning framework makes the layers fit $\mathbf{H}(\mathbf{x}) - \mathbf{x}$. In this case, if the input feature $\mathbf{x}$ contains all the useful information the residual learning framework will push $\mathbf{F}(\mathbf{x})$ to zero. By this mechanism, the deep residual network can be optimized easily and trained end-to-end.

\section{Introduction of Stacked-Hourglass Network}

\begin{figure}[h!]
  \centering
  \includegraphics[width=1\textwidth]{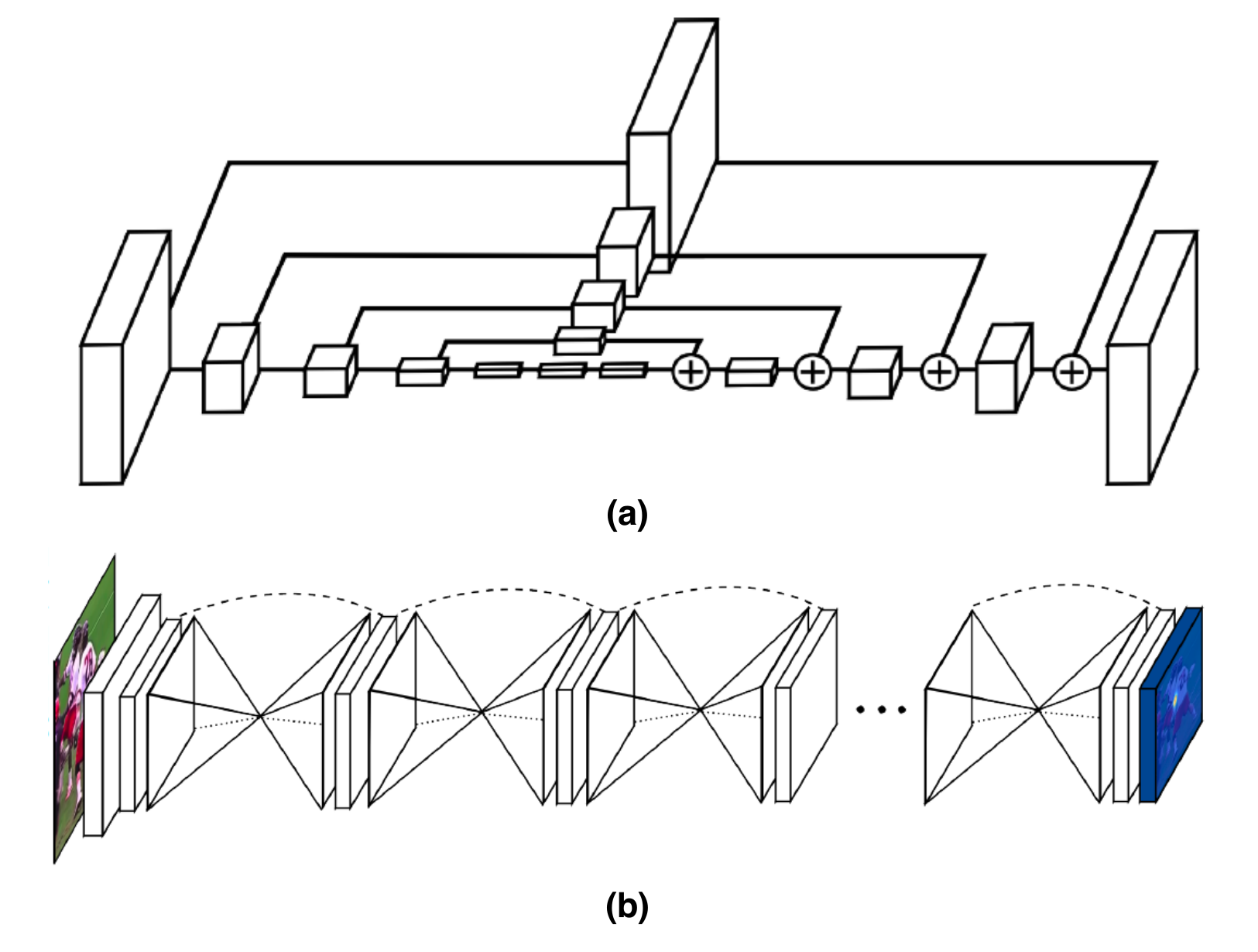}
  \vspace{-25pt}
  \caption{The figures come from \cite{newell2016stacked}. (a) is the single Hourglass module and (b) is the whole Stacked-Hourglass Network.}
  \label{fig:hourglass}
\end{figure}

The Stacked-Hourglass Network\cite{newell2016stacked} is designed for predicting human key points. The single Hourglass module in \figref{fig:hourglass} (a) is composed of many residual modules. The whole network in \figref{fig:hourglass} (b) consists of multiple Hourglass modules. These stacked Hourglass modules repeatedly downsample and upsample the features and catch the human pose information across various scales. During the training step, it predicts multiple heatmaps and applies the L2 loss on each of them. For the inference, it only keeps the heatmap which is produced by the last Hourglass module.

\chapter{Related Work}

\section{CNN based detection on natural images}
In recent years, deep convolutional neural network(CNN) based methods have been working with promising performance on natural images processing such as image classification \cite{krizhevsky2012imagenet}\cite{lecun1989backpropagation}\cite{wang2021contrastive}\cite{lanchantin2021general}, instance segmentation \cite{zhang2021refinemask}\cite{ke2021deep}\cite{ghiasi2021simple}\cite{wu2021dannet}, and object detection \cite{zhu2015segdeepm}\cite{sermanet2013overfeat}\cite{he2015spatial}\cite{redmon2016you}\cite{shafiee2017fast} tasks.  For the existing CNN based object detection methods, most of them can be classified into two groups: The regressing bounding-box based methods \cite{liu2016ssd}\cite{ren2015faster}\cite{tian2019fcos}\cite{qi2021multi} that regress the upper-left and lower-right corners of the bounding-boxes corresponding to the objects and the segmentation based methods \cite{newell2016stacked}\cite{zhang2021towards} that classify the pixels in the input image into different groups.

\textbf{Regressing bounding-box based methods.}
The regressing bounding-box based methods detect the objects by regressing the upper-left and lower-right coordinates of the bounding-boxes corresponding to the objects. For instance, as discussed in \cite{girshick2014rich}, it first generates the potential bounding-boxes based on the region proposals and then makes the classification and refines the bounding-boxes. To enhance the speed of detection, \cite{girshick2015fast} and \cite{ren2015faster} make an improvement on the proposal stage by computing the proposals with a deep convolutional neural network. In order to further increase the detection speed, \cite{redmon2016you} and \cite{liu2016ssd} propose the one-stage detection which spatially separates bounding-boxes and associates with class probabilities. Different from \cite{redmon2016you}, \cite{liu2016ssd} applied multiple aspect ratios and scales over different feature maps. Unlike \cite{redmon2016you}\cite{liu2016ssd},  \cite{tian2019fcos}\cite{qi2021multi} regress the location of the bounding-boxes in the image level directly instead of the feature level. Regarding the flexibility of detected bounding-boxes, \cite{ma2018arbitrary} suggests generating inclined proposals with angle information. Then it adjusts the angle of predicted bounding-boxes to make the detected result more accurate. However, these methods mentioned above focus more on the edges of the objects’ bounding boxes instead of the center point locations.  As a result, the calculated center point locations based on the upper-left and lower-right corners of the bounding-boxes are not precise.

\textbf{Segmentation based methods.}
The segmentation based methods classify the pixels in the input image into different groups and each group corresponds to one category. Different from the regressing bounding-box based methods, the segmentation based methods can detect the center point locations of the objects directly. For example, for the human key points detection task, the pixels in the input image are classified into background, mouth, nose, and so on. In \cite{newell2016stacked}, it concatenates multiple UNet\cite{ronneberger2015u} structures together that repeatedly downsample and upsample the features and catch the human key points information at different scales. To reduce the model parameter and speed up the detection, \cite{zhang2021towards} pay attention to the context information and propose a method called Cascaded Context Mixer which can integrate the spatial level and channel level information together and refine them step by step. 

\section{CNN based detection and segmentation on the material science images}
\textbf{Detection.}
In order to improve the detection speed and accuracy, in recent years, many works are trying to solve the detection problem in the material science field by using the deep learning based method. For example, \cite{yuyan2019internal} proposed to use Faster R-CNN based method to detect the internal defects from the CT scanning image of the metal three-dimensional lattice structure. To prevent overfitting to a small training dataset, it reduces the number of convolution layers and pooling layers. In \cite{napoletano2018anomaly}, a region based approach is implemented by CNN to detect and localize the anomalies in the scanning electron microscope images of nanofibrous materials. It also uses CNN-extracted features to evaluate the degree of difference between the anomaly samples and anomaly-free samples. In \cite{saeed2019automatic}, CNN and a Deep Feed Forward Network are combined to detect the anomaly of Carbon Fiber Reinforced Polymer thermograms. It can detect the anomaly in thermograms in real-time without any manual intervention. To speed up the detection and reduce the model parameters, \cite{chen2020light} proposed a method called WDD-Net which applies depthwise separable convolution and global pooling to detect the wafer structural defects. Besides the detection of anomalies and defects, in \cite{zhou2016large}, CNN is applied to detect the fiber cross-section in the 2D microscopic images, and the detected 2D fiber cross-sections are then used to reconstruct the 3D fiber structure.

\textbf{Segmentation.} Besides detection, CNN is also widely used for the segmentation of various material science images. For example, \cite{ma2018deep} proposed to train a CNN model based on DeepLab to perform the segmentation of AL-LA alloy microscopic images. It takes advantage of the local symmetric information and applies symmetric rectification to enhance the segmentation accuracy. In \cite{frei2021fiber}, FibeR-CNN is proposed to analyze the fiber images based on the segmentation result,  which employs R-CNN as the backbone and inserts additional convolution layers to help the prediction. To avoid labeling a large number of ground truth samples manually, \cite{chen2021semi} designed a semi-supervised learning method that applies UNet as the backbone to solve the segmentation problem of aluminum alloy metallographic image. This semi-supervised method only requires labeling a small number of images and can get better segmentation results than the traditional segmentation methods. To further improve the segmentation performance, \cite{panda2019deep} proposed the use of the Generative Adversarial Network (GAN) based method to perform the segmentation on carbon steel microstructure images. To reduce the number of training images, it also proposed a CNN based framework to generate the training data.

Detecting the dendrite core is a very challenging and important task that can help material scientists, there are very few prior researchs that are exactly focused on this problem. Most of the above CNN based detection methods for the materials science images treat all the detection targets equally without distinguishing the easy and hard samples. Besides, without considering the unique and complex properties of dendrites, general CNN based detection methods for the natural images may not work well for this specific task. In this work, we formulate the dendrite core detection problem as a segmentation task and propose a novel detection method to detect the easy samples and hard samples of dendrite cores separately.

\chapter{Methodology}

\begin{figure*}
\centering
\includegraphics[width=1.0\linewidth]{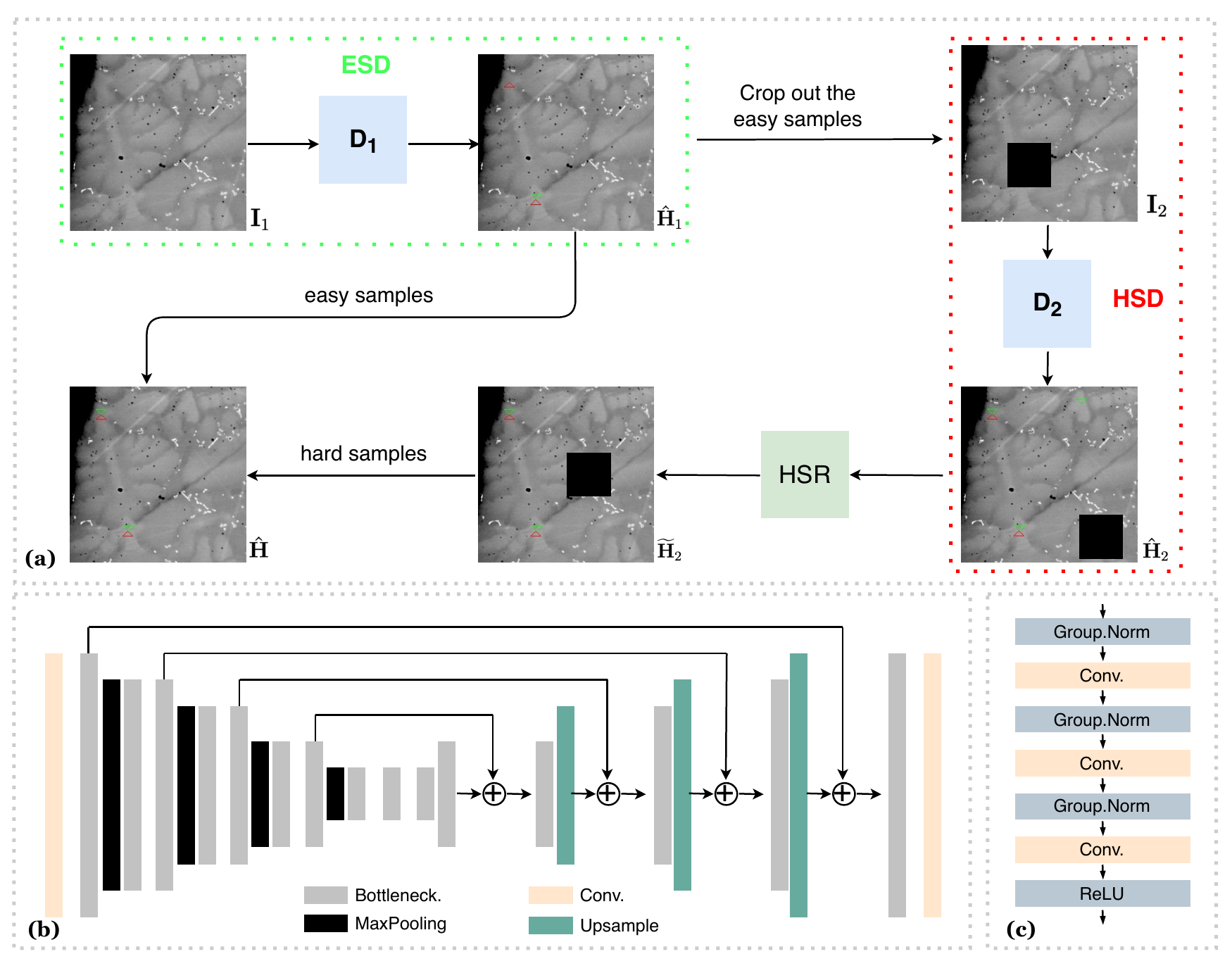}
\vspace{-40pt}
\caption{(a) denotes our whole pipeline, (b) denotes the Central Point Detection Network (CPDN), and (c) denotes the Bottleneck Block. Both $\mathbf{D}_1$ and $\mathbf{D}_2$ employ CPDN but not sharing parameters.
}
\label{fig:pipline}
\vspace{-10pt}
\end{figure*}

In this work, we formulate the dendrite core detection problem as a segmentation task. The whole pipeline contains three steps: Easy Sample Detection (ESD), Hard Sample Detection(HSD), and Hard Sample Refinement (HSR). Specifically, ESD and HSD focus on the easy samples and hard samples of dendrite cores respectively. Both of them employ the same Central Point Detection Network (CPDN), which is shown in \figref{fig:pipline} (b), but not sharing parameters. To make it clear, we denote the CPDN $\mathbf{D}_1$ in ESD and $\mathbf{D}_2$ in HSD. HSR is used to improve the Precision of HSD. The whole pipeline is shown in \figref{fig:pipline} (a). We will elaborate on ESD in \secref{subsec:esd}, HSD in \secref{subsec:hsd}, CPDN in \secref{subsec:dcdn}, and HSR in \secref{subsec:hsr} respectively.

\section{Easy Sample Detection (ESD)}
\label{subsec:esd}

In this step, it only focuses on detecting the easy samples of dendrite cores from the input image. We denote the input image $\mathbf{I}_1$ and ground truth heatmap $\mathbf{H}_1$ in ESD where $\mathbf{I}_1\in\mathds{R}^{H\times W \times 3}$ and $\mathbf{H}_1\in\mathds{R}^{H\times W \times 1}$. The ground truth heatmap $\mathbf{H}_1$ is binarized that one denotes the dendrite core and zero denotes the background. $\mathbf{D}_1$ will produce a heatmap $\hat{\mathbf{H}}_1$ to denote the detected location of each dendrite core based on the input image $\mathbf{I}_1$
\begin{align}\label{eq:o1_0}
\hat{\mathbf{H}}_1 = \mathbf{D}_1(\mathbf{I}_1)
\end{align}
where $\hat{\mathbf{H}}_1$ has the same dimension as $\mathbf{H}_1$. We aim to make ESD detect the easy samples of dendrite cores with a very high confidence to improve the Precision. The dendrite cores that can not be detected will be processed in HSD. Therefore, we set up a relatively higher confidence threshold $\alpha$ to binarize the predicted heatmap $\hat{\mathbf{H}}_1$. If the value of the location $(i, j)$ in $\hat{\mathbf{H}}_1$ greater than $\alpha$, we set $\hat{\mathbf{H}}_1(i, j)$ to one, otherwise set the value to zero, i.e., 
\begin{align}\label{eq:pre_H1}
\hat{\mathbf{H}}_1(i, j)= \left\{\begin{matrix}
 1,~\text{if}~\hat{\mathbf{H}}_1(i, j) \mathbf{>} \alpha,\\ 
0,\quad \quad \text{otherwise}.
\end{matrix}
\right.
\end{align}


\section{Hard Samples Detection (HSD)}
\label{subsec:hsd}

In this step, we only focus on the hard samples of dendrite cores which can not be detected in ESD such as the blurred or incomplete samples of dendrites. We denote the input image $\mathbf{I}_2$ and ground truth heatmap $\mathbf{H}_2$ in HSD  where $\mathbf{I}_2$ and $\mathbf{H}_2$ have same dimension as $\mathbf{I}_1$ and $\mathbf{H}_1$. 
To make $\mathbf{D}_2$ only focus on the feature of hard samples of dendrite cores, $\mathbf{I}_2$ shall only contain the hard samples of dendrites. We can construct $\mathbf{I}_2$ by cropping the structure of easy samples of dendrites which are detected by $\mathbf{D}_1$ from $\mathbf{I}_1$, i.e.,
\begin{align}\label{eq:I2}
\mathbf{I}_1(i-s:i+s, j-s:j+s) = 0, \quad \quad \quad \quad ~\text{if}~  \hat{\mathbf{H}}_1(i, j) = 1
\end{align}
where $s$ decide the crop size which will be discussed in \secref{subsec:crop}. The ground truth $\mathbf{H}_2$ is generated by:
\begin{align}\label{eq:g2}
\mathbf{H}_2(i, j)= \left\{\begin{matrix}
\quad 0, \quad \quad \quad \quad   ~\text{if}~\hat{\mathbf{H}}_1(i, j) \mathbf{=} \mathbf{1},\\ 
\mathbf{H}_1(i, j),\quad \quad \quad \quad \text{otherwise}.
\end{matrix}
\right.
\end{align}
$\mathbf{D}_2$ will produce another heatmap $\hat{\mathbf{H}}_2$ which denotes the detected locations of hard samples of dendrite cores based on $\mathbf{I}_2$, i.e.,
\begin{align}\label{eq:o1_0}
\hat{\mathbf{H}}_2 = \mathbf{D}_2(\mathbf{I}_2)
\end{align}
In HSD, it usually detects the hard samples of dendrite cores with a relatively lower confidence score. In order to improve the Recall, we setup a relative lower threshold $\beta$ to binarize the predicted heatmap $\hat{\mathbf{H}}_2$
\begin{align}\label{eq:pre_h2}
\hat{\mathbf{H}}_2(i, j)= \left\{\begin{matrix}
 1,~\text{if}~\hat{\mathbf{H}}_2(i, j) \mathbf{>} \beta,\\ 
0,\quad \quad \text{otherwise}.
\end{matrix}
\right.
\end{align}

\section{Central Point detection network (CPDN)}
\label{subsec:dcdn}
Both $\mathbf{D}_1$ and $\mathbf{D}_2$ employ the same CPDN but not sharing parameters. CPDN is implemented based on the architecture of U-Net. The encoder of CPDN consists of a convolution layer followed by several bottleneck layers and max-pooling layers. The bottleneck block consists of three group normalization layers and three convolution layers with kernel size 1*1, 3*3, and 1*1 respectively, followed by a ReLU operation as shown in \figref{fig:pipline} (c). The decoder of CPDN consists of several upsample layers and bottleneck layers, followed by a convolution layer with kernel size 1*1. $\mathbf{D}_1$ and $\mathbf{D}_2$ are trained separately and optimized with $L_2$ loss.

\section{Hard Samples Refinement (HSR)}
\label{subsec:hsr}

In order to improve the Recall in HSD, we set up a relative lower confidence threshold $\beta$ which leads to predict a lot of false positive samples of dendrite cores. To solve this problem, we add HSR after HSD. The HSR is a binary classifier which is gotten by fine tuning the ResNet-50 pretrained model on the dendrite dataset. We denote the input of HSR $\mathbf{I}_3$ which is a small patch with size 80*80 and can be gotten by
\begin{align}\label{eq:I3}
\mathbf{I}_3^{(i, j)} = \mathbf{I}_2(i-40:i+40, j-40:j+40), \quad \quad \quad ~\text{if}~ \hat{\mathbf{H}}_2(i, j) = 1
\end{align}
HSR will output one for the input $\mathbf{I}_3^{(i, j)}$ if $\hat{\mathbf{H}}_2(i, j)$ is a true positive prediction, otherwise output zero. We denote $\widetilde{\mathbf{H}}_2$ to be the refined result of $\hat{\mathbf{H}}_2$ and can get $\widetilde{\mathbf{H}_2}$ by
\begin{align}\label{eq:pre_h2}
\widetilde{\mathbf{H}}_2(i, j)= \left\{\begin{matrix}
 1,\quad~\text{if}~\mathbf{HSR}(\mathbf{I}_3^{(i, j)}) \mathbf{=} 1,\\ 
0,\quad \quad \quad \quad \text{otherwise}.
\end{matrix}
\right.
\end{align}
The effectiveness of HSR will be discussed later in \secref{chapter:ablation}. Finally, we get the predicted heatmap $\hat{\mathbf{H}}$ by
\begin{align}\label{eq:o}
\hat{\mathbf{H}}(i, j)= \left\{\begin{matrix}
1,~\text{if}~\hat{\mathbf{H}}_1(i, j) \mathbf{=} 1 ~\text{or}~ \widetilde{\mathbf{H}}_2(i, j) \mathbf{=} 1,\\ 
0,\quad \quad \quad \quad  \quad \quad  \quad \quad  \text{otherwise}.
\end{matrix}
\right.
\end{align}

\begin{figure}[h!]
  \centering
  \includegraphics[width=0.8\textwidth]{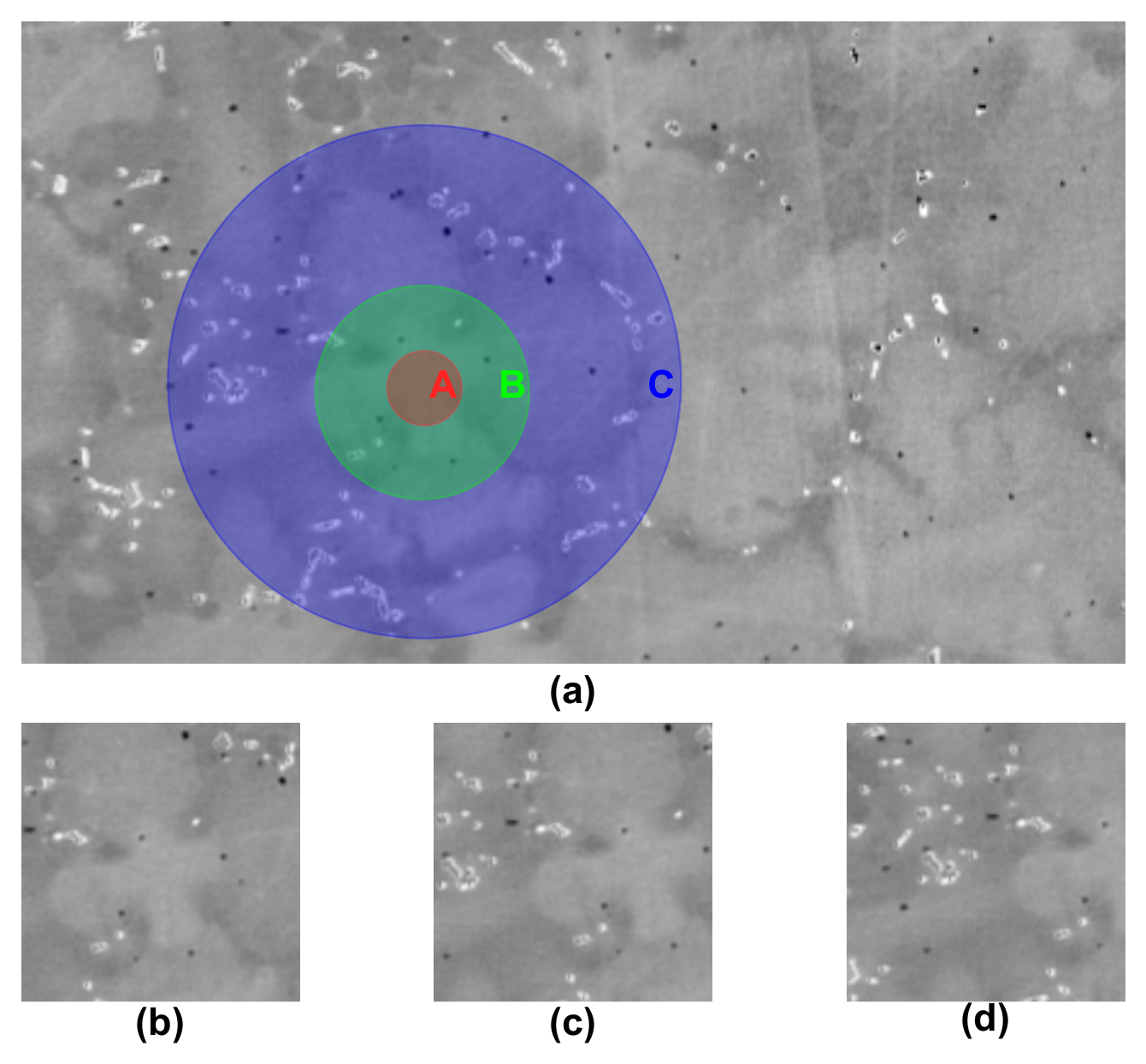}
  \vspace{-25pt}
  \caption{Building the training set for HSR. The positive samples are cropped around the randomly selected points inside circle $\mathbf{A}$, the negative samples are cropped around the randomly selected points inside the green area of circle $\mathbf{B}$ and the blue area of circle $\mathbf{C}$. The circles $\mathbf{A}$, $\mathbf{B}$, and $\mathbf{C}$ are around the dendrite core with radius 4, 15, 40 pixels respectively. (b) is an example of the positive sample and (c), (d) are the examples of negative samples.}
  \label{fig:bcn_sample}
\end{figure}

\textbf{Building the training dataset for HSR.} We randomly crop the 80*80 patches from the original images to build the training dataset for HSD. Specifically, around each dendrite core in the original images, we crop five positive samples and ten negative samples of dendrites. For the positive samples, first we draw a circle $\mathbf{A}$ as shown in \figref{fig:bcn_sample} around each of dendrite cores with the radius 4 pixels. Then we randomly choose five locations inside circle $\mathbf{A}$ and crop a patch with the size 80*80 around each of the five locations. It is the same for the negative samples, we draw another two circles $\mathbf{B}$, $\mathbf{C}$ as shown in \figref{fig:bcn_sample} around each of dendrite cores with the radius 15 and 40 pixels respectively. Then we randomly choose five locations inside the green area of circle $\mathbf{B}$, blue area of circle $\mathbf{C}$ respectively and crop a patch with the size 80*80 around each of the locations. We show a positive sample in \figref{fig:bcn_sample} (b) and two negative samples in \figref{fig:bcn_sample} (c), (d).

\section{Loss Function}
\label{subsec:loss}
We use the $L_2$ loss as objective function to train $\mathbf{D}_1$ and $\mathbf{D}_2$ separately which is shown as:
\begin{align}\label{eq:loss}
\mathcal{L}_1 ~\text{=}~ \lambda_1 * ||\hat{\mathbf{H}}_1 - \mathbf{H}_1||_2
\end{align}
\begin{align}\label{eq:loss}
\mathcal{L}_2 ~\text{=}~ \lambda_2 * ||\hat{\mathbf{H}}_2 - \mathbf{H}_2||_2
\end{align}



\section{Implementation details}

We train $\mathbf{D}_1$ in ESD with a learning rate of 0.0004.  Then, we train $\mathbf{D}_2$ in HSD with a learning rate of 0.0001 by fixing the parameter of $\mathbf{D}_1$. We train both $\mathbf{D}_1$ and $\mathbf{D}_2$ 100 epochs. Then, we train HSR 50 epochs with a learning rate of 0.0001. We set up the threshold $\alpha$ to 0.4,  $\beta$ to 0.1, and $\lambda_1$ and $\lambda_2$ in the loss function to 0.5. The experiments are conducted on the same platform with two NVIDIA Tesla V100 GPUs.


\chapter{Experiments}
\section{Setups}
\label{subsec:setups}

\textbf{Dataset.} The dendrites dataset includes 1,461 microscopic images with resolution 512*512. We divide the dataset into a training set (60\%), a validation set (20\%), and a test set (20\%). 

\textbf{Metrics.} We follow the previous object detection methods and use Recall, Precision, and F-score to evaluate our method.

\textbf{Baselines.} We compare our method with three state-of-the-art object detection algorithms: Faster R-CNN, SSD, and Stacked-Hourglass.


\section{Comparison Results}

\begin{figure}[h!]
\centering
  \includegraphics[width=0.98\textwidth]{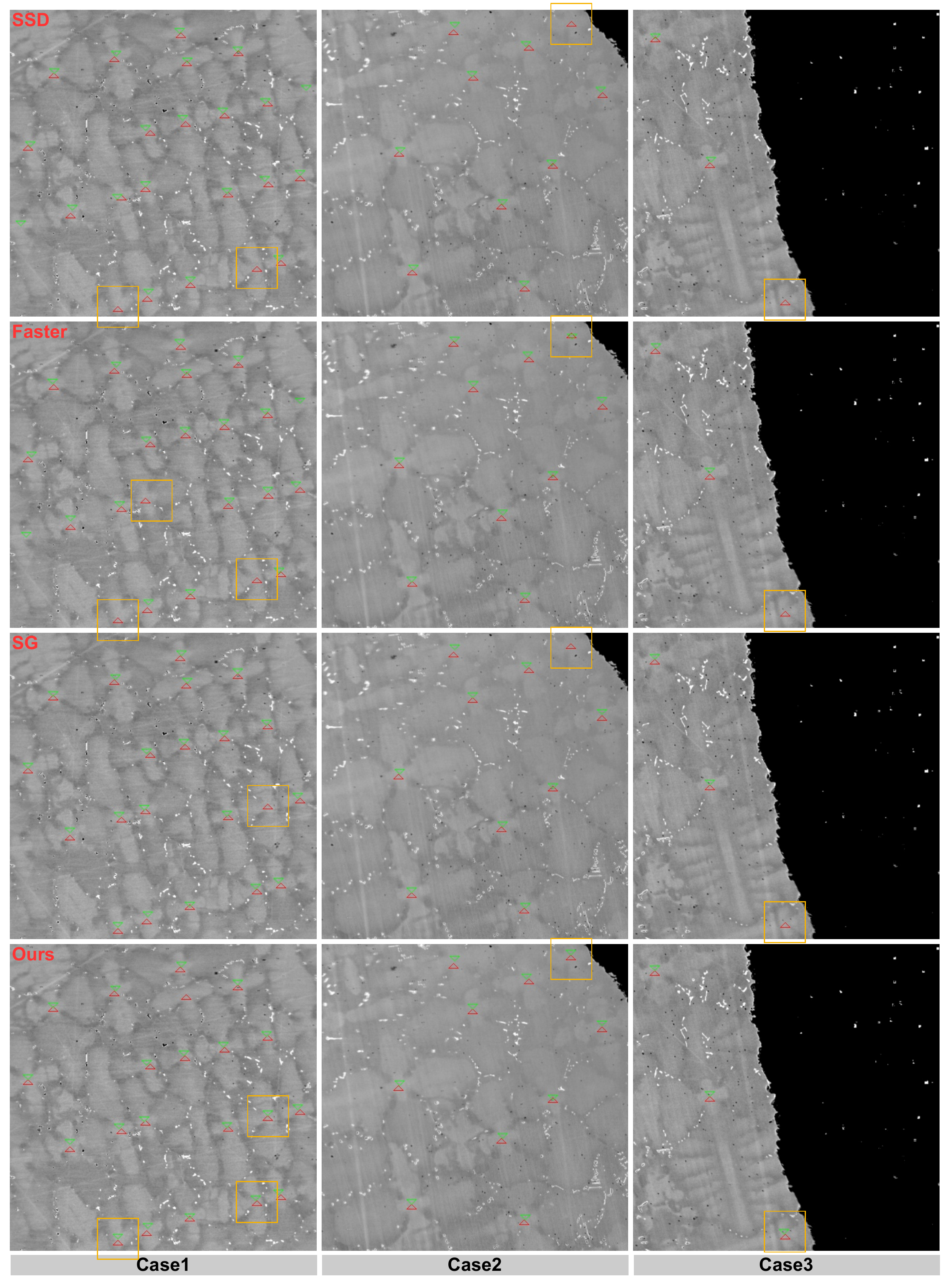}
  \vspace{-25pt}
\caption{Three visualization results of \cite{liu2016ssd}, \cite{ren2015faster}, \cite{newell2016stacked}, and our method. The upper points of the red triangles and the lower points of green triangles denote the ground truth and predicted dendrite cores respectively. We highlight the hard samples of dendrite cores detected by our method by the yellow boxes.}
  \label{fig:comparison_result}
\end{figure}

\subsection{Quantitative Results}

We compare our method with three state-of-the-art object detection methods and show the comparison results in \tableref{tab:comparison} and \tableref{tab:comparison_number}. From \tableref{tab:comparison} and \tableref{tab:comparison_number}, we can observe that with the deviation distance(the distance between the predicted location and ground truth) of 10 pixels, our method can reach the best result on Recall, Precision, and F-score. Specifically, comparing to the regressing bounding-box based detection \cite{ren2015faster}, our method can achieve 5.77\% relative higher Recall, 7.29\% relative higher Precision, and 6.53\% relative higher F-score and comparing to the segmentation based detection \cite{newell2016stacked}, our method can achieve 0.77\% relative higher Recall, 1.58\% relative higher Precision, and 1.18\% relative higher F-score.

\subsection{Qualitative Results}

We display the visualization results on three samples taken from the dendrite dataset in \figref{fig:comparison_result}. From the visualization result, we can find that our method has the ability to detect more hard samples of dendrite cores such as the blurred or incomplete dendrites inside the yellow boxes, while the other methods can not.

\begin{table*}
\centering
\small
\caption{Comparison results of \cite{liu2016ssd}, \cite{ren2015faster}, \cite{newell2016stacked}, and our method on Recall, Precision, and F-Score.}
{
    \resizebox{1.0\linewidth}{!}{
    {
	\begin{tabular}{l|l|c|c|c}
		
    \toprule
     \multicolumn{1}{l|}{Methods} & \multicolumn{1}{l|}{Deviation} & \multicolumn{1}{c|}{Recall$\uparrow$} & \multicolumn{1}{c|}{Precision$\uparrow$}  & \multicolumn{1}{c}{F-Score$\uparrow$}  \\
   \midrule
   \multirow{1}{*}{SSD}                
      & 10  & 0.9585 & 0.9415 & 0.9499\\
      
    \midrule
    \multirow{1}{*}{Faster-RCNN}   
    & 10    & 0.9112 & 0.9050 & 0.9081\\

    \midrule
    \multirow{1}{*}{Stacked-Hourglass}   
     & 10   &  0.9564 & 0.9559 & 0.9561\\
     
    \midrule
    \multirow{1}{*}{\topone{Our}}   
     & \topone{10}  & \topone{0.9638} & \topone{0.9710} & \topone{0.9674}\\

    \bottomrule
		
	\end{tabular}
	}
	}
}
\label{tab:comparison}
\end{table*}



\begin{table*}
\centering
\small
\caption{Comparison results of \cite{liu2016ssd}, \cite{ren2015faster}, \cite{newell2016stacked}, and our method based on the number of dendrite cores detected.}
\label{tab:comparison_number}
{   
    \resizebox{1.0\linewidth}{!}{
    {
	\begin{tabular}{l|l|c|c|c}
		
    \toprule
     \multicolumn{1}{l|}{Methods} & \multicolumn{1}{l|}{Deviation} & \multicolumn{1}{c|}{Total} & \multicolumn{1}{c|}{T-Positive$\uparrow$}  & \multicolumn{1}{c}{F-Positive$\downarrow$}  \\
   \midrule
   \multirow{1}{*}{SSD}                
      & 10  & 1882 & 1804 & 112\\
      
    \midrule
    \multirow{1}{*}{Faster-RCNN}   
    & 10    & 1882 & 1715 & 180\\

    \midrule
    \multirow{1}{*}{Stacked-Hourglass}   
     & 10   &  1882 & 1800 & 83\\
     
    \midrule
    \multirow{1}{*}{\topone{Ours}}   
     & 10  & 1882 & \topone{1814} & \topone{54}\\

    \bottomrule
		
	\end{tabular}
	}
	}
}

\end{table*}


\section{Ablation Study}
\label{chapter:ablation}

\subsection{Quantitative Results}

To validate the effectiveness of our method, we consider three variants: ESD (\secref{subsec:esd}), HSD (\secref{subsec:hsd}), and HSR (\secref{subsec:hsr}). From \tableref{tab:ablation} and \tableref{tab:ablation_number}, we can observe that: \ding{182} Only using ESD, it gets a lower Recall which leads to a worse F-score. \ding{183} The HSD is effective to increase the Recall. Specifically, it can increase 3.52\% of Recall compared to ESD. \ding{184} HSR is effective to filter out the false positive prediction of HSD and increase Precision. Specifically, it can increase 1.54\% of Precision compared to the HSD. As a result, by taking advantages of both HSD and HSR, our method achieves the best F-score.

\subsection{Qualitative Results}

\begin{figure}[h!]
  \includegraphics[width=1\textwidth]{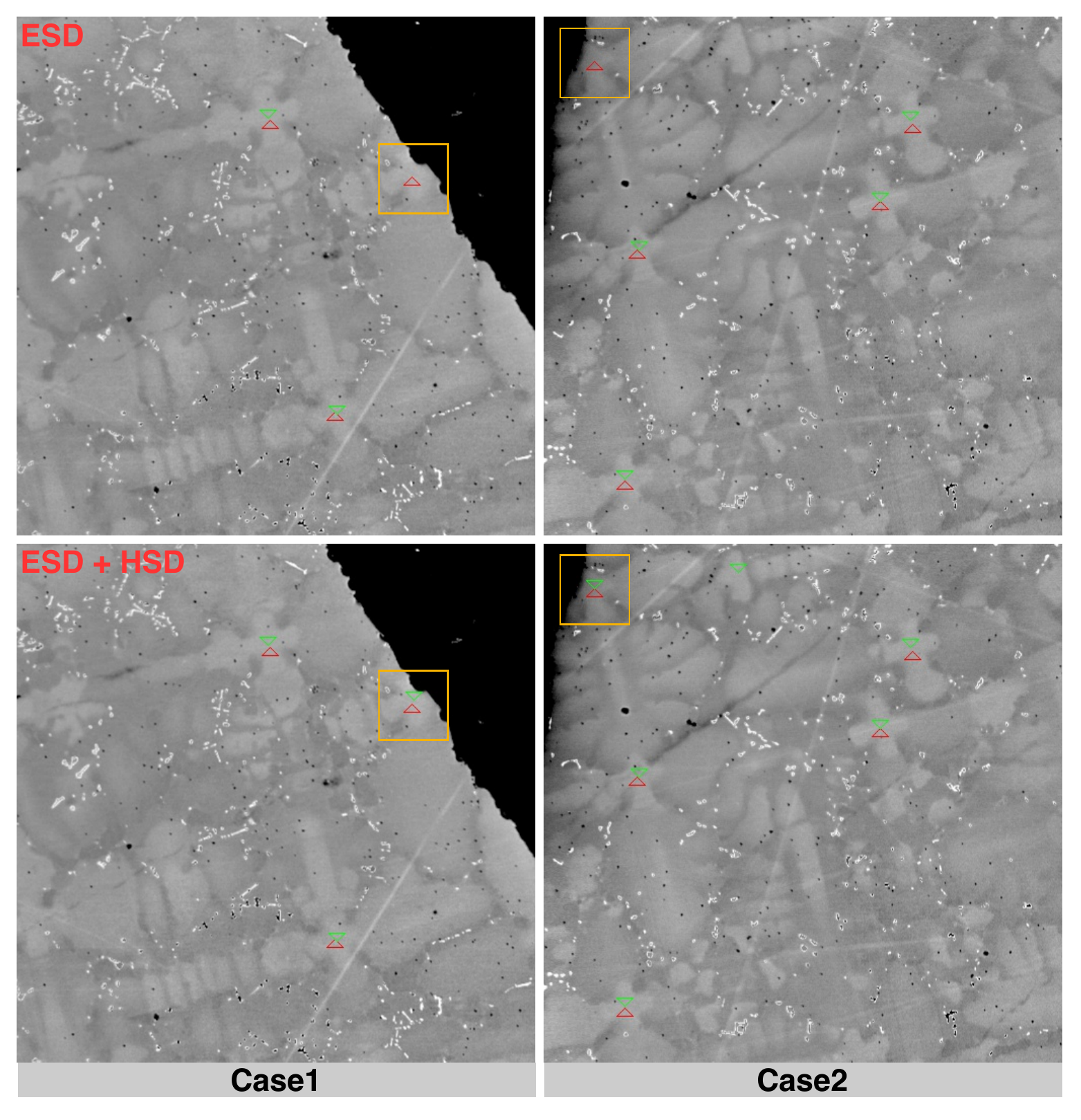}
  \vspace{-25pt}
  \caption{Two visualization results of the ESD and ESD + HSD. The upper points of the red triangles and the lower points of green triangles denote the ground truth and the predicted dendrite cores respectively. The yellow boxes denote the hard samples of dendrite cores detected by the HSD.}
  \label{fig:ablation_esdn_hsdn}
\end{figure}

\begin{figure}[h!]
  \includegraphics[width=1\textwidth]{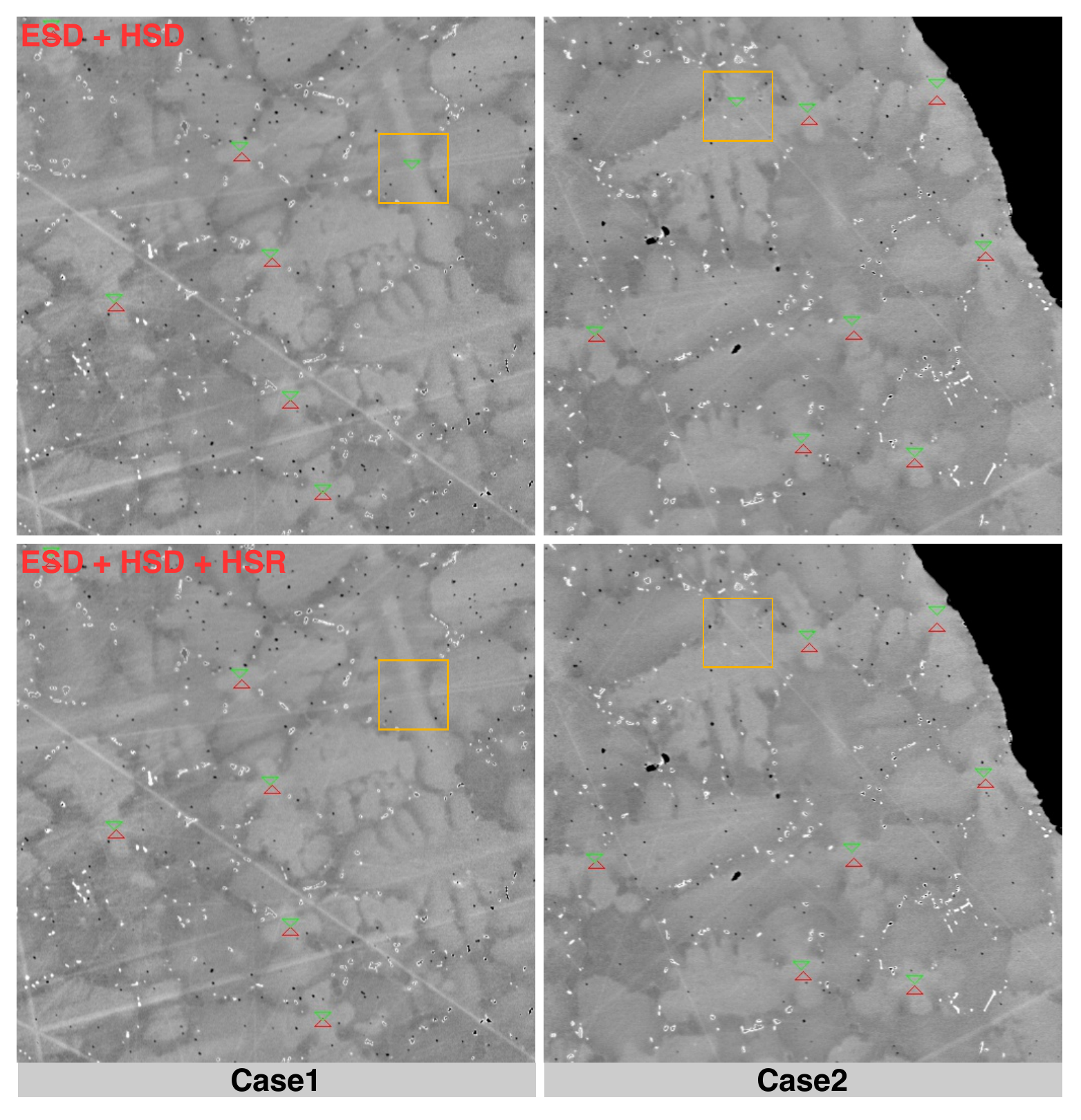}
  \vspace{-25pt}
  \caption{Two visualization results of the ESD + HSD and ESD + HSD + HSR. The upper points of the red triangles and the lower points of green triangles denote the ground truth and the predicted dendrite cores respectively. The yellow boxes denote the false positive predictions filtered out by the HSR.}
  \label{fig:ablation_hsdn_bcn}
\end{figure}

We show the visualization result in \figref{fig:ablation_esdn_hsdn} and \figref{fig:ablation_hsdn_bcn} to prove the effectiveness of each component. From the visualization result, we can find that \ding{182} HSD has the ability to detect more hard samples of dendrite cores such as the dendrite cores inside the yellow boxes in \figref{fig:ablation_esdn_hsdn}. \ding{183} HSR can filter out the false-positive detection such as the dendrite cores inside the yellow boxes in \figref{fig:ablation_hsdn_bcn}. As a result, ESD, HSD, and HSR can work together to increase the F-score.

\begin{table*}
\centering
\small
\caption{Ablation study on ESD, HSD, and HSR. We show the result for ESD in the first row, ESD + HSD in the second row and ESD + HSD + HSR in the third row. In the ablation study, we use the deviation distance 10 pixels.}
{
    \resizebox{0.8\linewidth}{!}{
    {
	\begin{tabular}{l|l|c|c|c}
		
    \toprule
     \multicolumn{1}{l|}{Methods} & \multicolumn{1}{l|}{Deviation} & \multicolumn{1}{c|}{Recall$\uparrow$} & \multicolumn{1}{c|}{Precision$\uparrow$}  & \multicolumn{1}{c}{F-Score$\uparrow$}  \\
   \midrule
   \multirow{1}{*}{ESD}                
      & 10  & 0.9341 & 0.9745 & 0.9538\\
      
    \midrule
    \multirow{1}{*}{+HSD}   
    & 10    & 0.9670 & 0.9563 & 0.9616\\

    \midrule
    \multirow{1}{*}{\topone{+HSR}}   
     & \topone{10}  & \topone{0.9638} & \topone{0.9710} & \topone{0.9674}\\

    \bottomrule
		
	\end{tabular}
	}
	}
}

\label{tab:ablation}
\end{table*}

\begin{table*}
\centering
\small
\caption{Ablation study on ESD, HSD, and HSR based on the number of dendrite cores detected. We show the result for ESD at the first row, ESD + HSD at the second row and ESD + HSD + HSR at the third row. In the ablation study, we use the deviation distance 10 pixels.}
{
    \resizebox{0.8\linewidth}{!}{
    {
	\begin{tabular}{l|l|c|c|c}
		
    \toprule
     \multicolumn{1}{l|}{Methods} & \multicolumn{1}{l|}{Deviation} & \multicolumn{1}{c|}{Total} & \multicolumn{1}{c|}{T-Positive$\uparrow$}  & \multicolumn{1}{c}{F-Positive$\downarrow$}  \\
   \midrule
   \multirow{1}{*}{ESD}                
      & 10  & 1882 & 1800 & 83\\
      
    \midrule
    \multirow{1}{*}{+HSD}   
    & 10    & 1882 & \topone{1820} & 83\\

    \midrule
    \multirow{1}{*}{\topone{+HSR}}   
     & 10  & 1882 &  1814 & \topone{54}\\

    \bottomrule
		
	\end{tabular}
	}
	}
}

\label{tab:ablation_number}
\end{table*}

\section{Exploiting the crop size}
\label{subsec:crop}

\begin{figure}[h!]
  \includegraphics[width=1\textwidth]{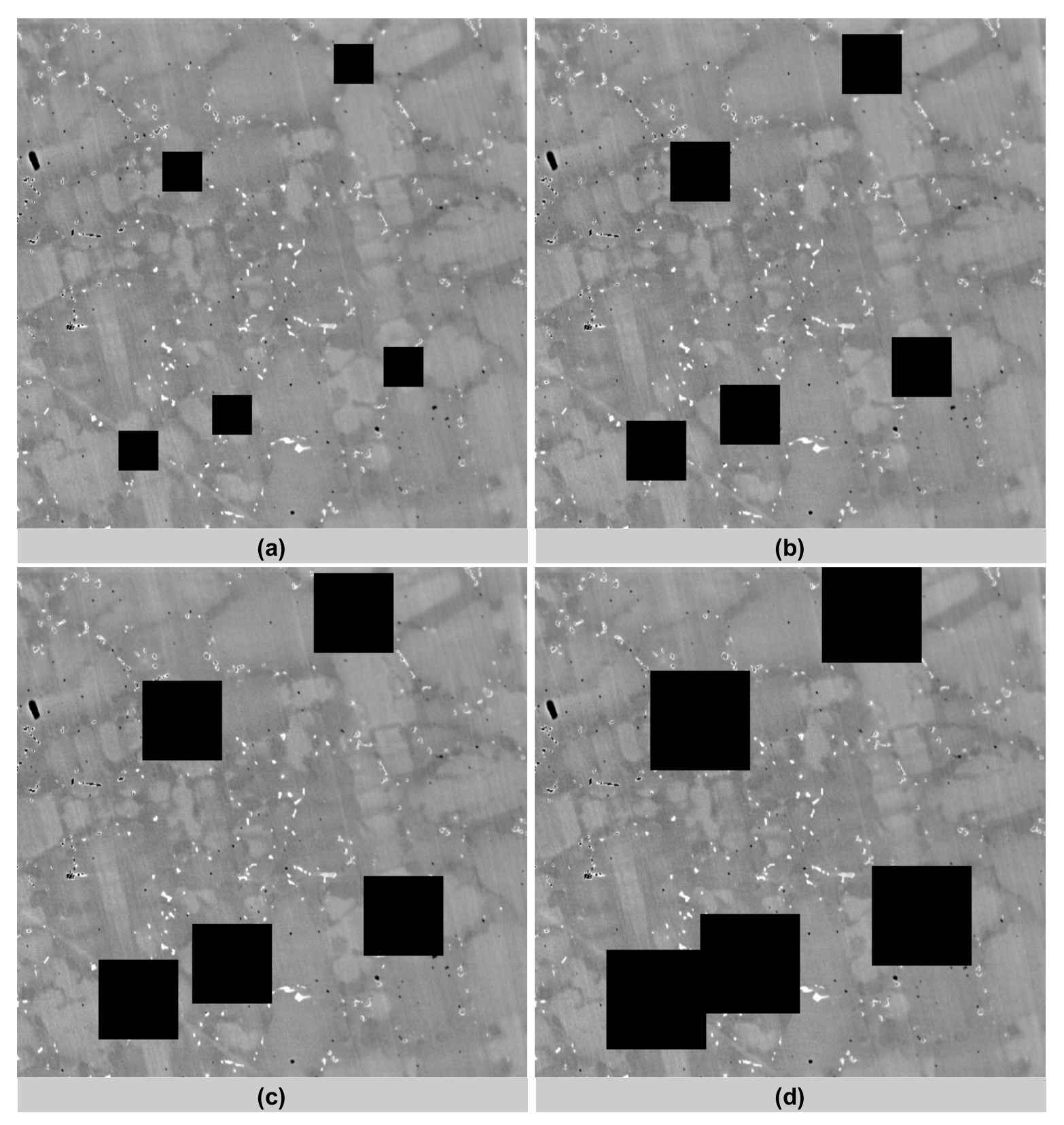}
  \vspace{-25pt}
  \caption{The visualization of different crop sizes. (a) denotes the crop size 40*40, (b) denotes the crop size 60*60, (c) denotes the crop size 80*80, and (d) denotes the crop size 100*100.}
  \label{fig:crop_size}
\end{figure}

During the procedure of destroying the structure of easy samples of dendrites detected by ESD, the crop size is an important factor that can affect the final detected result. In order to exploit the reasonable crop size, we have conducted a series of experiments. We take different crop sizes to destroy the structure of easy samples of dendrites. Specifically, we use the crop size 40*40, 60*60, 80*80, 100*100, and no crop, respectively. The visualization results of the cropped images are displayed in \figref{fig:crop_size}. 

\begin{figure}[h!]
  \includegraphics[width=1\textwidth]{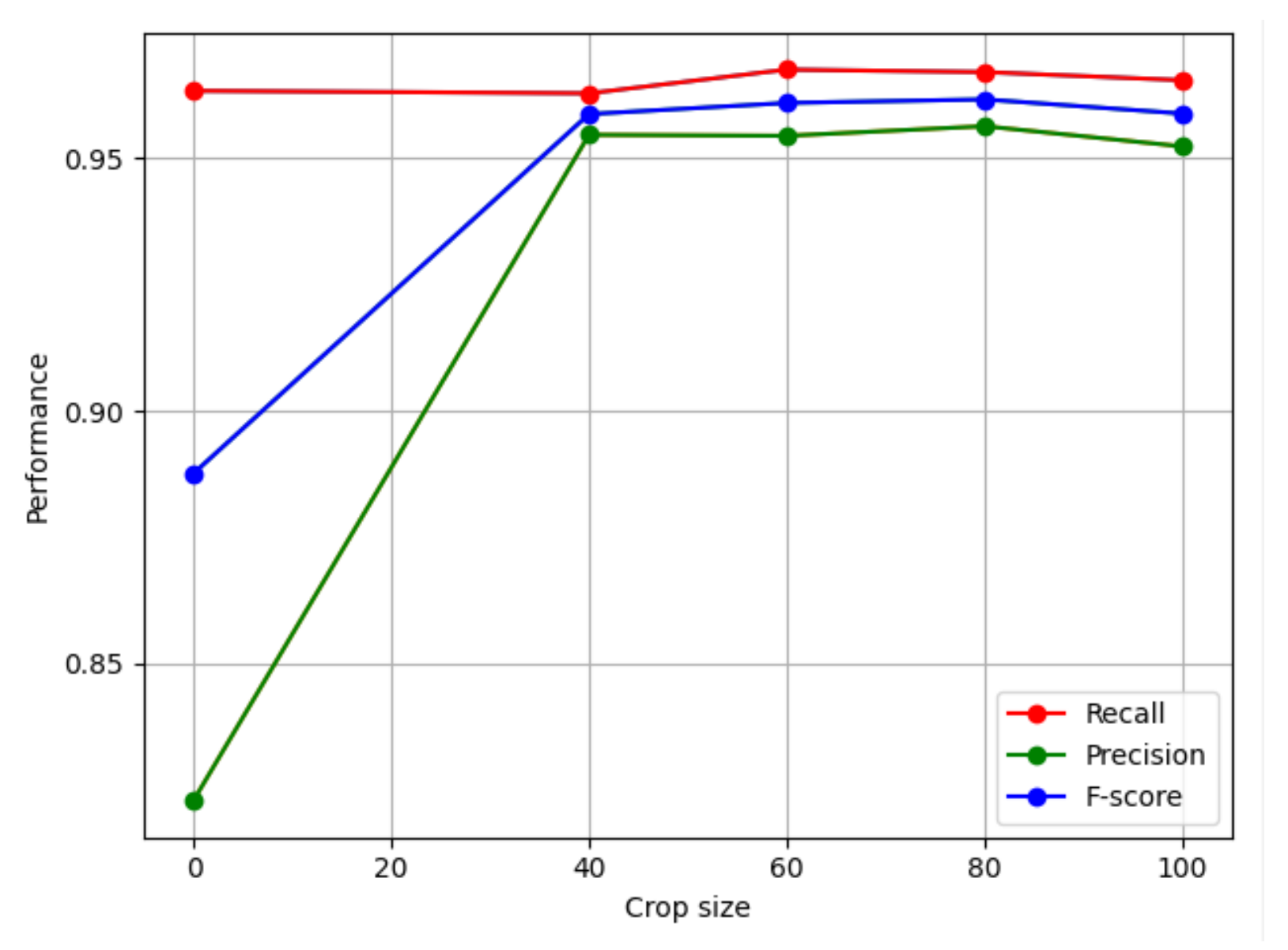}
  \vspace{-25pt}
  \caption{Exploiting the optimal crop size after ESD. Red points denote the Recall, green points denote the Precision, and blue points denote the F-score. When the crop size is equal to 80*80, it gets the best F-score.}
  \label{fig:crop_curve}
\end{figure}

We evaluate the Recall, Precision, and F-score on the dendrite dataset which is discussed in \secref{subsec:setups} with different crop sizes and show the results in \tableref{tab:crop_size} and \figref{fig:crop_curve}. From \figref{fig:crop_curve}, we can see that \ding{182} Without destroying the structure of easy samples of dendrites detected by ESD, the final F-score is very low. \ding{183} It reaches the best F-score when the crop size is 80*80. When continuing to increase the crop size to 100*100, the F-score decreases. The reason for this result is that the smaller crop size or without cropping can not destroy the structure of dendrites detected by ESD and the larger crop size may destroy the potential dendrites which should be detected by HSD. Based on these observations, we finally choose the crop size 80*80.

\begin{table*}
\centering
\small
\caption{Exploiting the optimal crop size after ESD when the deviation distance is 10 pixels. In this work, we try the crop size: 40*40, 60*60, 80*80, 100*100, and no crop, respectively.}
{
    \resizebox{0.8\linewidth}{!}{
    {
	\begin{tabular}{l|c|c|c}
		
    \toprule
     \multicolumn{1}{l|}{Crop size} & \multicolumn{1}{c|}{Recall$\uparrow$} & \multicolumn{1}{c|}{Precision$\uparrow$}  & \multicolumn{1}{c}{F-Score$\uparrow$}  \\
     
   \midrule
                 
       no crop  & 0.9633 & 0.8229 & 0.8876\\
   
   \midrule
                
       40*40  & 0.9628 & 0.9546 & 0.9587\\
 
    \midrule
                
       60*60  & 0.9675 & 0.9544 & 0.9609\\

    \midrule
  
      \topone{80*80}  & \topone{0.9670} & \topone{0.9563} & \topone{0.9616}\\

    \midrule
              
       100*100  & 0.9654 & 0.9523 & 0.9588\\

    \bottomrule
		
	\end{tabular}
	}
	}
}

\label{tab:crop_size}
\end{table*}



\section{Exploiting the intensity}
\label{subsec:intensity}

\begin{table*}
\centering
\vspace{-15pt}
\small
\caption{Exploiting the optimal intensity in the cropped areas when deviation distance is 10 pixels and cropped size is 80*80. 'Gaussian' means destroy the structure of easy samples by only using Gaussian smoothing. '0 + Gaussian' means first filling the cropped areas with intensity 0 and then processing the cropped boundary by Gaussian smoothing.}
{
    \resizebox{0.8\linewidth}{!}{
    {
	\begin{tabular}{l|c|c|c}
		
    \toprule
      \multicolumn{1}{l|}{Intensity} & \multicolumn{1}{c|}{Recall$\uparrow$} & \multicolumn{1}{c|}{Precision$\uparrow$}  & \multicolumn{1}{c}{F-Score$\uparrow$}  \\
    
    \midrule
 
      \topone{0}  & \topone{0.9670} & \topone{0.9563} & \topone{0.9616}\\
   
   \midrule
             
       128  & 0.9659 & 0.9478 & 0.9568\\

    \midrule
             
       255  & 0.9675 & 0.9549 & 0.9612\\

   \midrule
             
       Gaussian & 0.9617 & 0.9597 & 0.9607\\ 

    \midrule
            
       0 + Gaussian  & 0.9617 & 0.9541 & 0.9579\\

    \bottomrule
		
	\end{tabular}
	}
	}
}

\label{tab:intensity}
\end{table*}

\subsection{Filling the cropped area with fixed intensity}
\label{subsec:intensity_fixe}

\begin{figure}[h!]
  \includegraphics[width=1\textwidth]{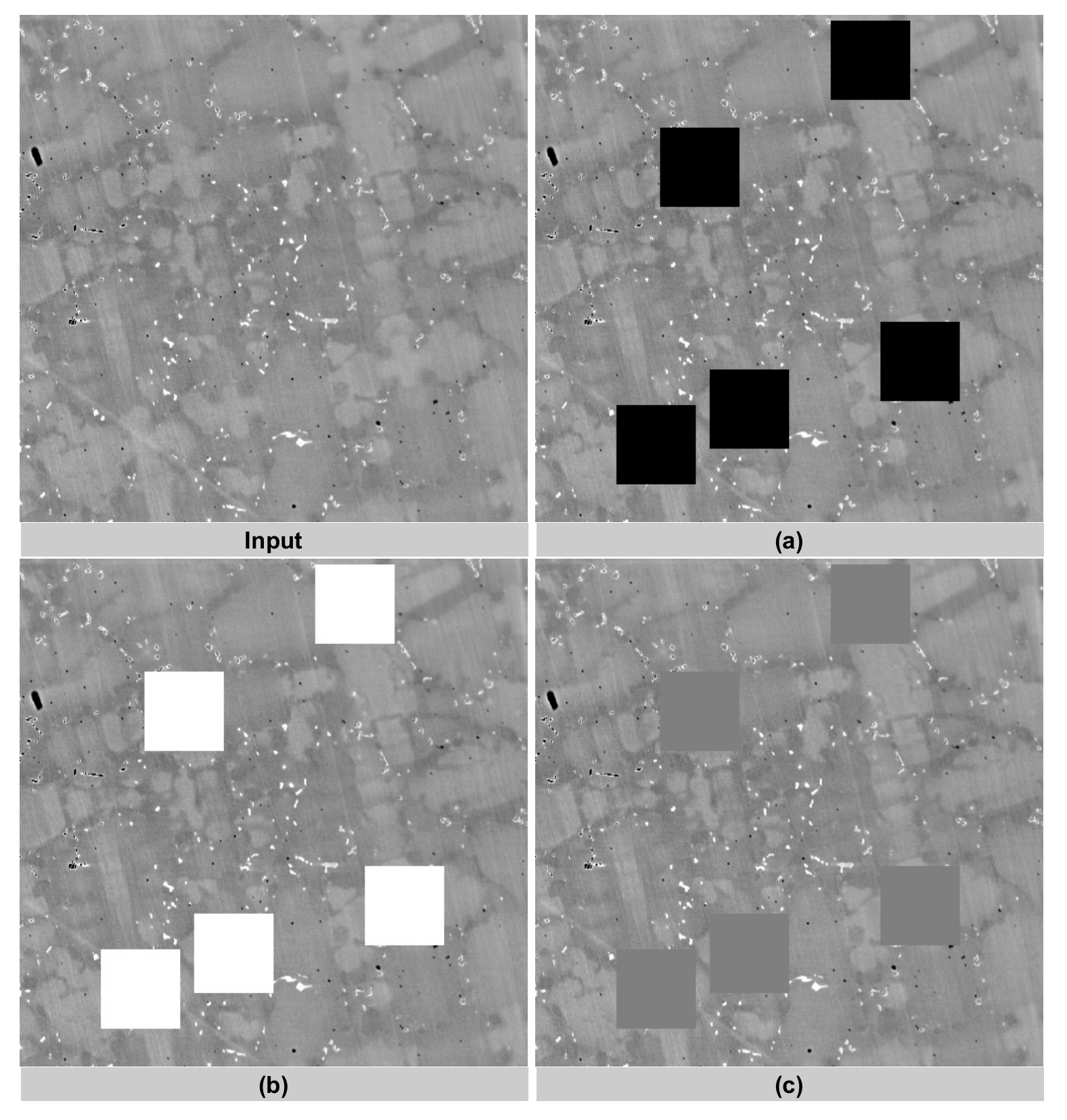}
  \vspace{-25pt}
  \caption{Exploiting the optimal intensity to fill the cropped area. The intensity in the cropped areas is 0 for (a), 255 for (b), and 128 for (c).}
  \label{fig:crop_intensity}
\end{figure}

\begin{figure}[h!]
  \includegraphics[width=1\textwidth]{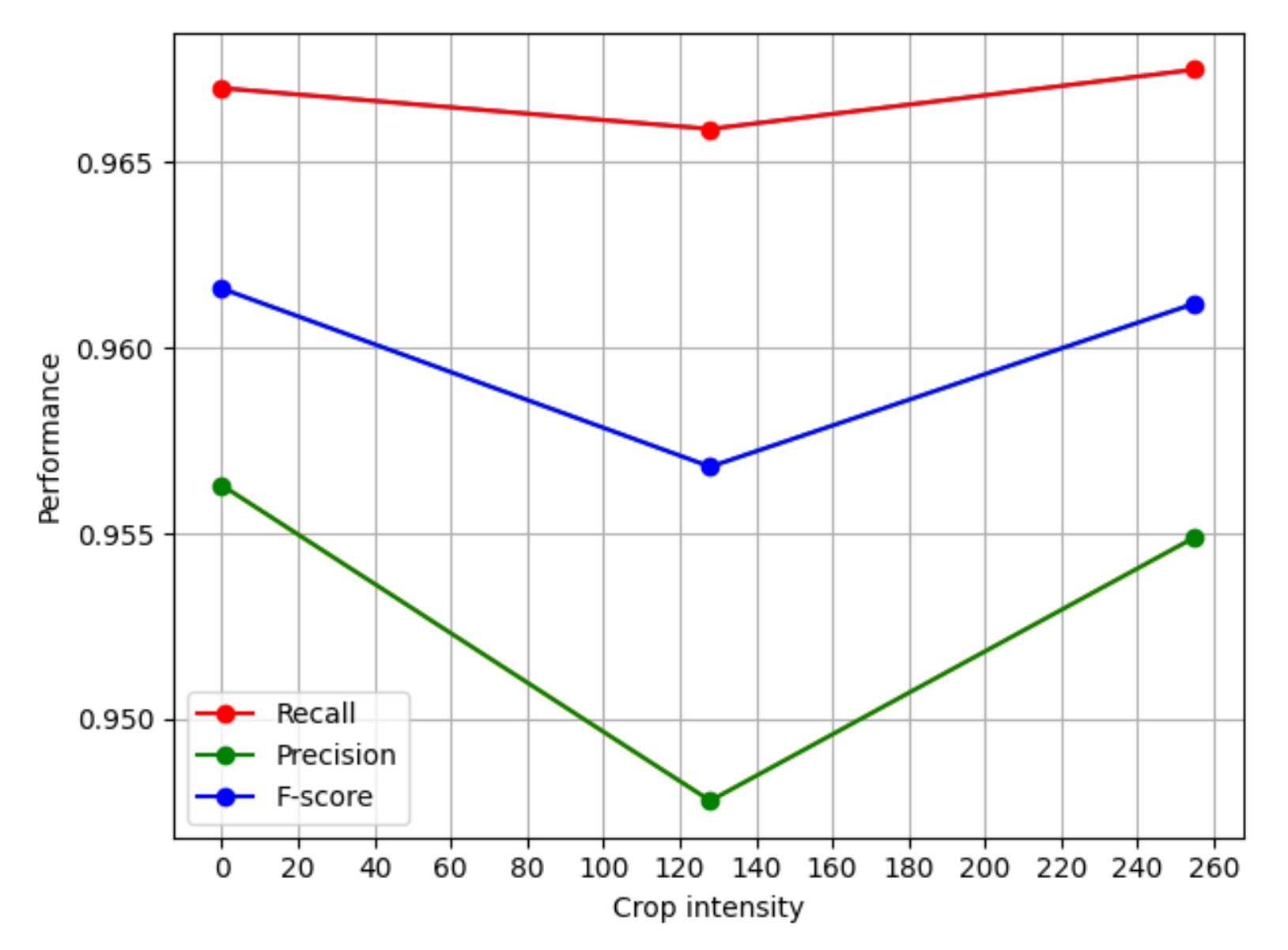}
  \vspace{-25pt}
\caption{Exploiting the optimal intensity to fill the cropped area. In this work, we try three intensities 0, 128, and 255, respectively. Red points denote the Recall, green points denote the Precision, and blue points denote the F-score. When the intensity is equal to 0, it gets the best F-score.}
  \label{fig:intensity_curve}
\end{figure}

In addition to the crop size, the crop intensity also has a big influence on the prediction result of HSD. In this work, we conduct a series of experiments to exploit the optimal crop intensity on dendrite dataset. Specifically, we try the fixed intensities of 0, 128, and 255, respectively. Based on the discussion in \secref{subsec:crop}, using crop size 80*80 can reach the best F-score. Therefore, we only consider the crop size 80*80 here. The visualization results of different crop intensity are shown in \figref{fig:crop_intensity}. We also evaluate the Recall, Precision, and F-score with the different intensities. From \figref{fig:intensity_curve} we find that with the intensity 0, it can get the best F-score. Finally, we choose the intensity 0 in this work.

\subsection{Processing the cropped area with Gaussian smoothing}

\begin{figure}[h!]
  \includegraphics[width=1\textwidth]{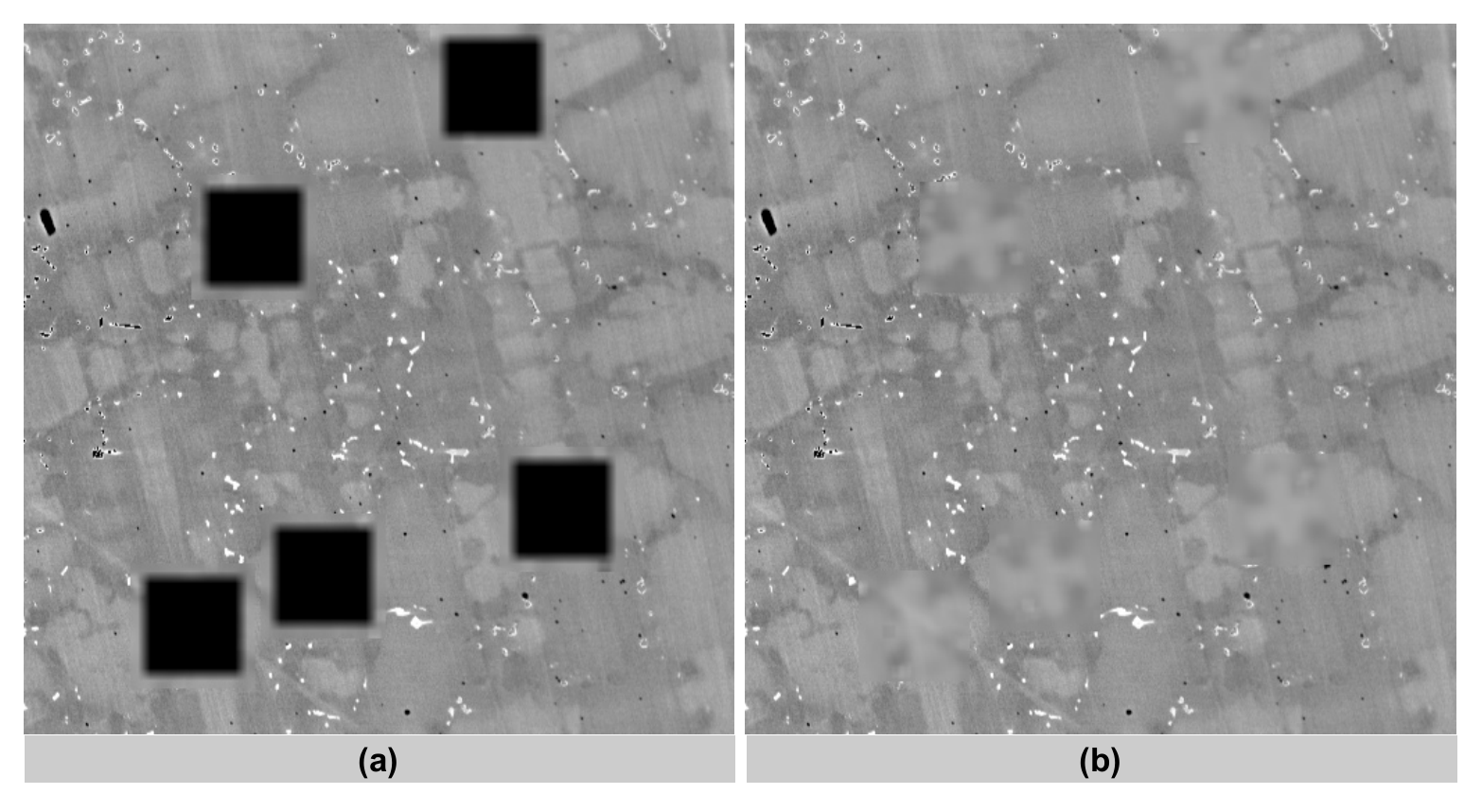}
  \vspace{-25pt}
\caption{Using Gaussian smoothing to process cropped areas. (a) denotes first to fill the cropped area with intensity 0 and then use Gaussian smoothing to process the edges. (b) Only using Gaussian smoothing to destroy the structure of easy samples of dendrites.}
  \label{fig:gaussian_sample}
\end{figure}

In addition to fill the cropped area with the fixed intensity, we also exploit using the Gaussian smoothing to process the cropped area. The standard deviation used to generate the kernel of Gaussian smoothing is in the range [1, 20] and the kernel size is 11*11. Based on the discussion in \secref{subsec:intensity_fixe}, using the intensity 0, the HSD can get the best F-score. Therefore, we use the Gaussian smoothing to process the boundaries of the cropped areas with intensity 0. The visualization results are shown in \figref{fig:gaussian_sample} (a). Besides, we also try to destroy the structure of easy samples by only using Gaussian smoothing. The visualization results are shown in \figref{fig:gaussian_sample} (b). From \tableref{tab:intensity} we can see that with the fixed intensity 0, it can get the best F-score.

\chapter{Conclusions}

In this work, we formulate the object detection problem as a segmentation task and proposed a novel detection method to detect the dendrite cores directly. The whole pipeline contains Easy Sample Detection, Hard Sample Detection, and Hard Sample Refinement. The Easy Sample Detection and Hard Sample Detection focus on the easy samples and hard samples of dendrite cores respectively. Both of them employ the same Central Point Detection Network but not sharing parameters. The Hard Sample Refinement is a binary classifier which is used to filter out the false positive prediction of Hard Sample Detection. We also conducted a series of experiments for exploiting a way to destroy the structure of easy samples of dendrites detected by Easy Sample Detection. As a result, our proposed detection method outperforms the state-of-the-art baselines on three metrics, i.e., Recall, Precision, and F-score. 

One potential limitation of our method is that we use different components to focus on improving the Recall and Precision separately which increases the difficulty to train the whole pipeline. E.g., we use Hard Sample Detection to increase the Recall and use Hard Sample Refinement to increase the Precision. In future work, we will exploit new approaches to combine all the components into a single network.

\printbibliography 

\Appendix                 


\end{document}